\documentclass[letterpaper]{article} 
\usepackage[preprint]{aaai2027}
\usepackage[hyphens]{url}  
\usepackage{graphicx} 
\usepackage{natbib}  
\usepackage{caption} 
\usepackage{algorithm}
\usepackage{algorithmic}
\usepackage{graphicx}
\usepackage{booktabs}
\usepackage{tabularx}
\usepackage{array}

\newcolumntype{Y}{>{\centering\arraybackslash}X}
\usepackage{amsmath}
\usepackage{amssymb}
\usepackage{dblfloatfix}

\usepackage{newfloat}
\usepackage{listings}
\DeclareCaptionStyle{ruled}{labelfont=normalfont,labelsep=colon,strut=off} 
\floatstyle{ruled}
\newfloat{listing}{tb}{lst}{}
\floatname{listing}{Listing}

\usepackage{booktabs}

\title{Beyond Final Decisions: A Process-Centric Benchmark for Transparent AI-Assisted Peer Review}
\author{
    Siming Yuan\textsuperscript{\rm 1}, Xueyi Zhang\textsuperscript{\rm 2}, Wangze Ni\textsuperscript{\rm 1}, Tianfang Xiao\textsuperscript{\rm 3}, Shimin Di\textsuperscript{\rm 4}, Jia Zhu\textsuperscript{\rm 5}, Zhuoren Jiang\textsuperscript{\rm 1}, Rong Tan\textsuperscript{\rm 1}, Lei Chen\textsuperscript{\rm 6}, Kui Ren\textsuperscript{\rm 1}\\
}
\affiliations{
    \textsuperscript{\rm 1}Zhejiang University \\

    \textsuperscript{\rm 2}Nanyang Technological University\\
    \textsuperscript{\rm 3}School of Business, Sun Yat-sen University\\
    \textsuperscript{\rm 4}Southeast University\\
    \textsuperscript{\rm 5}Zhejiang Normal University\\
    \textsuperscript{\rm 6}DSA, The Hong Kong University of Science and Technology
}

\begin{document}

\maketitle

\begin{abstract}
Peer review is central to quality control in science. However, existing evaluations of AI-assisted peer review mainly focus on the overall quality of generated reviews or the accuracy of final decisions. They therefore provide limited evidence about whether model decisions are supported by sufficient and reliable review evidence. We introduce a process-centric diagnostic benchmark for AI-assisted peer review. It uses \((x,z_s,z_c,z_r,y)\) to represent the paper content, summary, critique, suggestion, and decision. We convert heterogeneous review records from PeerRead, NLPeer ARR-22, and OpenReview-ICLR into process-aligned data. Our benchmark uses direct decision prediction from the paper content (Direct) as its baseline. It compares the decision value of Gold-process variables and Predicted-process variables, and conducts stage-level evaluation, chain-consistency evaluation, and interventional sensitivity analysis. Experiments across three datasets and six models show that Gold-process variables generally have higher decision value. For the main analysis model, the Gold--Predicted gap remains stable across datasets and random seeds. This gap is also reproduced in most model--dataset combinations. Although model-generated intermediate review texts show relatively high local consistency across adjacent stages, the final decisions are not consistently supported by the preceding review evidence. Our benchmark targets AI systems designed to assist rather than replace human reviewers. It provides a transparent and auditable diagnostic tool for evaluating the reliability of their review processes.
\end{abstract}
\section{Introduction}
As submission volumes to artificial intelligence conferences continue to grow, peer review faces an increasing workload. Review quality, timeliness, and accountability are also under growing pressure ~\cite{kim2025peerreviewcrisis}. Large language models have been used to generate review comments, assist with paper analysis, and predict review scores and acceptance decisions. Some conferences have also begun controlled pilot programs to explore their use in real review workflows ~\cite{aaai2025reviewpilot,neurips2026reviewexperiment}. However, the growing use of AI-assisted peer review has raised concerns about fairness and research integrity. Studies on specific conference data have found that AI-assisted reviews tend to assign higher scores and may affect the outcomes of papers near the acceptance threshold ~\cite{latona2024aireviewlottery}. Therefore, human reviewers must retain final decision authority and accountability, while transparent and auditable evaluation methods are needed to assess the reliability of AI-provided review assistance.
\begin{figure}[t]
\centering
\includegraphics[width=\columnwidth]{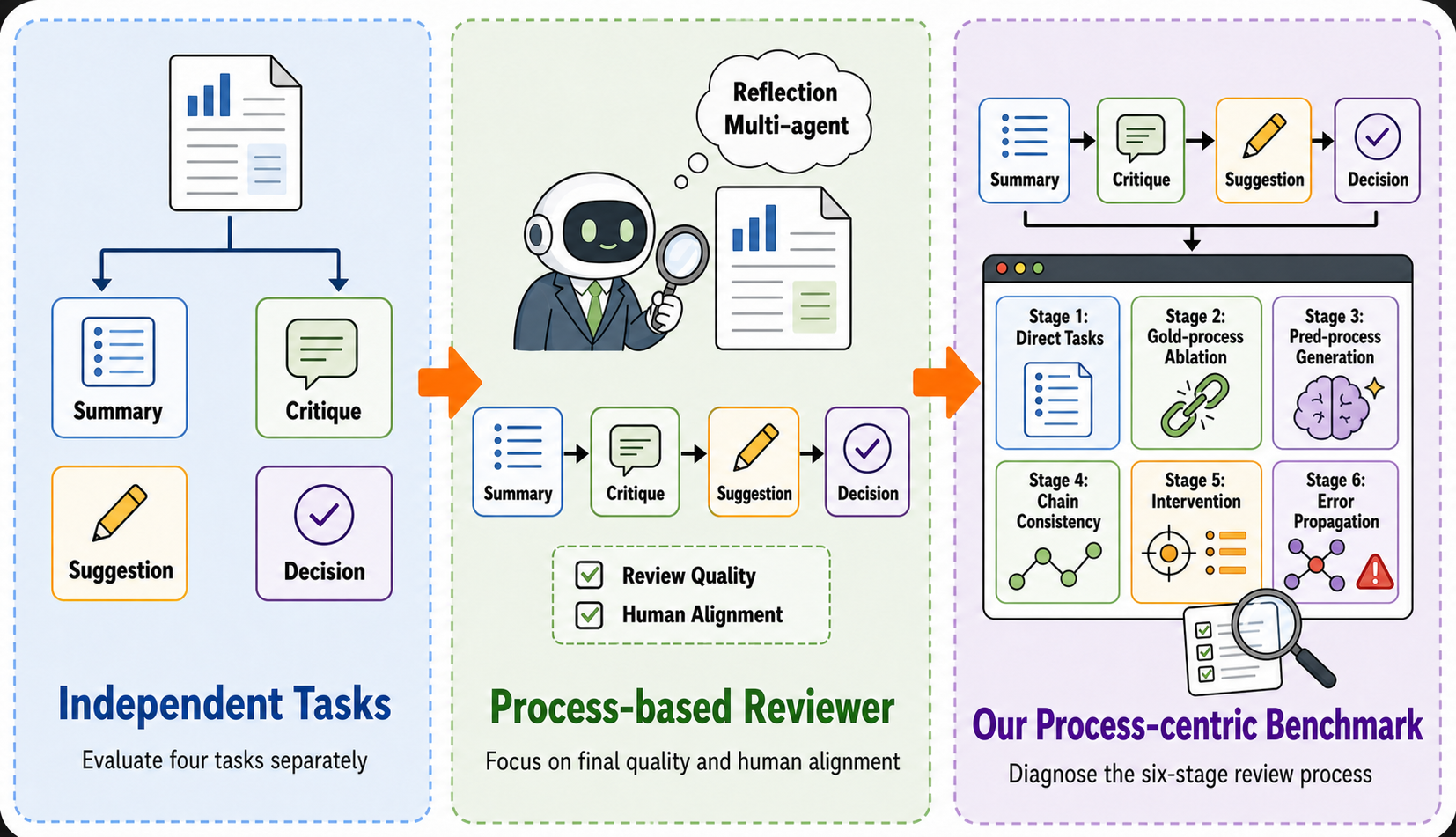}
\caption{From independent-task evaluation to process-centric benchmarking
for AI-assisted peer review.}
\label{fig:overview}
\end{figure}
Peer review is a process in which domain experts analyze a paper, identify its main contributions and problems, provide suggestions for improvement, and form an overall assessment~\cite{liang2023can,idahl2025openreviewer,biswas2026ai}. For example, the NeurIPS review form records these review components in separate fields ~\cite{neurips2025reviewerguidelines}. Inspired by this structure, we represent the core evidence-forming process of a single review as four stages: summary, critique, suggestion, and decision. As shown in Figure 1, this representation supports not only the evaluation of the functional quality of each stage, but also the assessment of chain consistency across stages and whether the final decision is sufficiently supported by the preceding review evidence.

Existing evaluations mainly focus on performance on independent tasks ~\cite{kang2018dataset,zhou2024llm} or on the final review quality and human alignment of process-based reviewers~\cite{d2024marg,weng2025cycleresearcher,zhu2025deepreview,garg2025revieweval}. Few studies compare process-variable decision value, evaluate chain consistency, and examine the interventional sensitivity of final decisions to intermediate process variables within a unified framework. Using a multi-stage or structured procedure to generate reviews does not mean that the intermediate stages and their relationships have been systematically evaluated. As AI-assisted peer review enters controlled real-world pilot programs, closing this evaluation gap is important for improving system transparency, clarifying accountability, and supporting reliable deployment.

However, process-level evaluation presents two key challenges. First, existing peer-review datasets usually contain heterogeneous paper--review--decision records. Their field structures, text granularity, and label sources differ, making it difficult to align summary, critique, and suggestion into unified process variables~\cite{kang2018dataset,dycke2023nlpeer,wang2023have}. Second, these process variables are open-ended texts with no single correct expression. An output that appears reasonable at one stage therefore does not establish consistency across the complete review chain~\cite{liang2023can,idahl2025openreviewer}. It also does not show that the final decision actually uses the corresponding evidence. 
Therefore, process-level evaluation must assess the functional quality of each stage and examine process-variable decision value, chain consistency, and the influence of process variables on final decisions.

To address these challenges, we introduce a process-centric diagnostic benchmark for AI-assisted peer review. To support stage-wise diagnosis of the core review components, we represent each review instance as \((x,z_s,z_c,z_r,y)\). These variables correspond to the paper content, summary, critique, suggestion, and decision. We then align heterogeneous review records from PeerRead ~\cite{kang2018dataset}, NLPeer ARR-22 ~\cite{dycke2023nlpeer}, and OpenReview-ICLR~\cite{idahl2025openreviewer} into unified process-aligned data.

Based on this representation, we first compare decision performance under three input settings: paper content only, Gold-process variables aligned from human review records, and Predicted-process variables generated by models. This comparison measures the process-variable decision value of the two types of process variables. We then conduct stage-level evaluation, chain-consistency evaluation, interventional sensitivity analysis, and conditional error analysis. Together, these evaluations diagnose how models generate and use review evidence.

The results show that Gold-process variables generally have higher decision value. For the main analysis model, Qwen2.5-14B-Instruct, the Gold--Predicted gap remains stable across datasets and random seeds. The gap is also reproduced in most model--dataset combinations. Although model-generated intermediate review texts show relatively high local consistency across adjacent stages, the final decisions are not consistently supported by the preceding review evidence. Therefore, final-label accuracy and overall review quality alone cannot fully evaluate the reliability of a model's review process. Our benchmark targets AI systems designed to assist rather than replace human reviewers. It evaluates the reliability of the review evidence provided by these systems and does not treat model outputs as a standalone basis for acceptance decisions.

The main contributions of this work are as follows:
\begin{itemize}
    \item We introduce a process-centric diagnostic benchmark for AI-assisted peer review. It uses \((x,z_s,z_c,z_r,y)\) to represent the paper content, summary, critique, suggestion, and decision. Based on this unified representation, we convert heterogeneous papers, human reviews, and decision records from PeerRead, NLPeer ARR-22, and OpenReview-ICLR into comparable process-aligned data.
    
    \item We develop a process-level evaluation framework that compares the decision value of Gold-process variables and Predicted-process variables. It also conducts stage-level evaluation, chain-consistency evaluation, interventional sensitivity analysis, and conditional error analysis to systematically diagnose how models generate and use review evidence.   
    \item We conduct systematic experiments across three datasets and six models. The results reveal a decision-value gap between Gold-process variables and Predicted-process variables. Although model-generated intermediate review texts show relatively high local consistency across adjacent stages, the final decisions are not consistently supported by the preceding review evidence.
\end{itemize}

\section{Related Work}
\textbf{AI-Assisted Peer Review.} AI-assisted peer review mainly covers review generation, score prediction, and acceptance prediction. Early work explored the automatic generation of initial review comments ~\cite{yuan2022can}, while PeerRead enabled computational analysis of review texts and acceptance decisions ~\cite{kang2018dataset}. More recent systems use hierarchical question answering, retrieval augmentation, iterative reflection, and multi-agent collaboration to build stronger reviewers. Representative systems include TreeReview ~\cite{chang2025treereview}, ScholarPeer ~\cite{goyal2026scholarpeer}, ReviewerTool~\cite{sahu2510reviewertoo}, and DeepReviewer~\cite{zhu2025deepreview}. These methods improve the depth, factuality, or structure of generated reviews, but their main goal remains improving the reviewer or its final output. In contrast, our work examines whether intermediate review evidence contains decision-relevant information, remains consistent across stages, and is reflected in the final decision.

\textbf{Process-Oriented Evaluation.} Evaluation has expanded from score or decision accuracy to multidimensional review quality and human alignment. Beyond Rating~\cite{li2026beyond}, ReviewEval~\cite{garg2025revieweval}, MMReview~\cite{gao2025mmreview}, PRISM~\cite{loc2026prism}, and CoCoReviewBench ~\cite{deng2026cocoreviewbench} evaluate generated reviews in terms of factuality, completeness, constructiveness, and alignment with human reviews. Concern-level matching further supports more fine-grained diagnosis~\cite{jin2026makes}. Existing data resources provide complementary foundations. PeerRead contains papers, reviews, and acceptance labels~\cite{kang2018dataset}, while NLPeer standardizes multiple peer-review data sources~\cite{dycke2023nlpeer}. PeerSum~\cite{li2022peersum}, MOPRD~\cite{lin2023moprd}, and resources for peer-review argumentation~\cite{fromm2021argument} support review summarization, multi-stage review records, and discourse analysis, respectively. However, existing evaluations usually focus on individual tasks or final review quality. Few align summary, critique, suggestion, and decision as unified and intervenable process variables. We provide such a unified representation and evaluate stage-level quality, process-variable decision value, chain consistency, interventional sensitivity, and conditional error associations.
\begin{figure*}[t]
\centering
\includegraphics[width=0.83\textwidth]{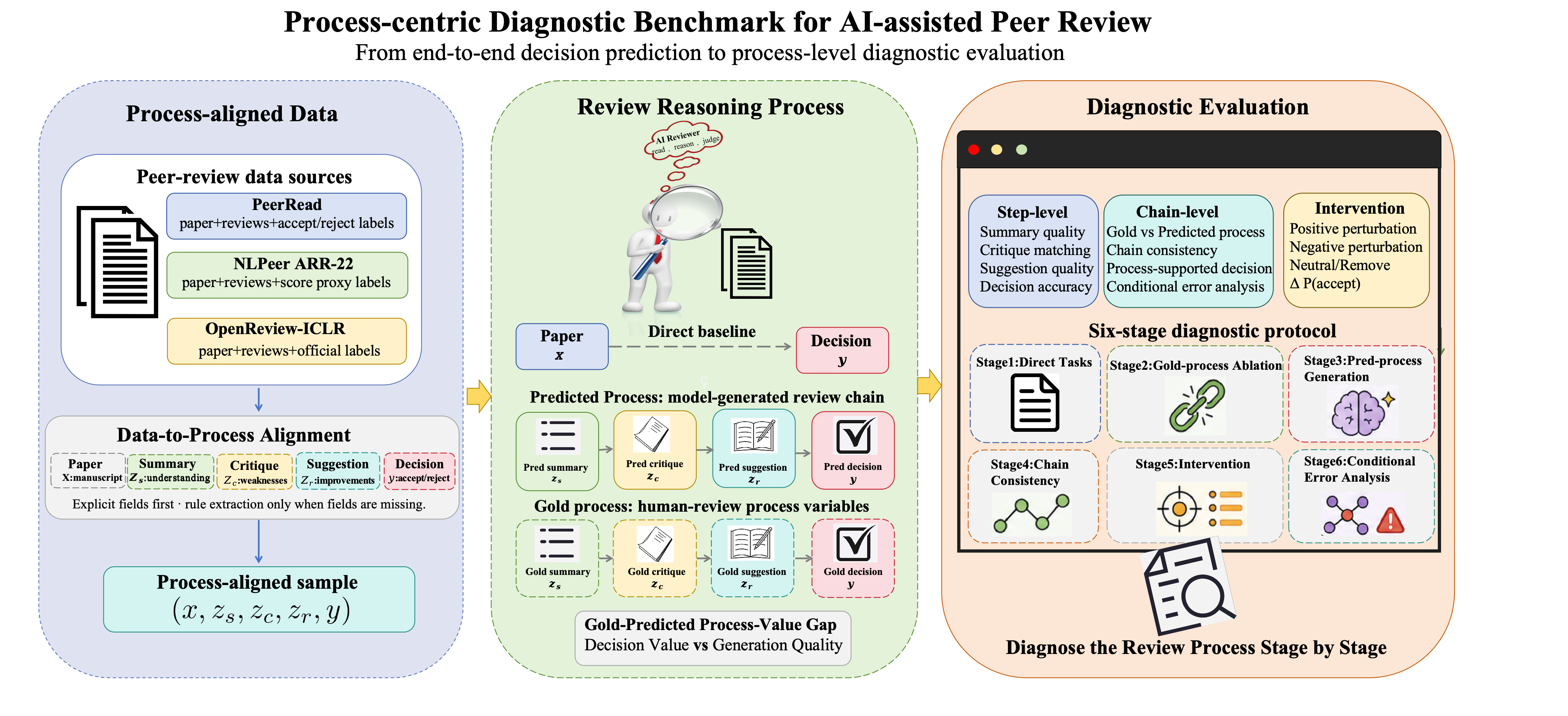} 
\caption{Overview of the proposed process-centric diagnostic benchmark for AI-assisted peer review. It aligns heterogeneous peer-review records into process variables and evaluates their decision value, stage quality, chain consistency, interventional sensitivity, and conditional error relations.}
\label{fig:framework}
\end{figure*} 
\section{Process-Centric Benchmark}
We introduce a process-centric diagnostic benchmark for AI-assisted peer review. As shown in Figure 2, the benchmark converts heterogeneous peer-review records into process-aligned data. It analyzes how models generate and use review evidence by comparing the decision value of Gold-process and Predicted-process variables and conducting process-level diagnostic evaluation.
\subsection{Problem Formulation}
Existing outcome-level evaluations typically model AI-assisted peer review as the direct prediction of a final decision \(y\) from paper content \(x\): $\hat{y}^{\mathrm{direct}}=f_{\mathrm{direct}}(x), y\in\{\mathrm{accept},\mathrm{reject}\}.$ This setting measures whether the final label is correct, but it cannot reveal whether the decision is supported by intermediate review evidence. To enable process-level diagnosis, we represent each review sample as \((x,z_s,z_c,z_r,y)\). Here, \(x\) denotes the paper content; \(z_s\), \(z_c\), and \(z_r\) denote the summary, critique, and suggestion; and \(y\) denotes the final decision. It enables separate evaluation, combination, and intervention of these stages. Gold-process variables are aligned from human review records and denoted as \(Z_G=(z_s,z_c,z_r)\). Predicted-process variables are generated by the evaluated model and denoted as \(\hat{Z}_P=(\hat{z}_s,\hat{z}_c,\hat{z}_r)\). Gold-process variables serve as reference representations for evaluation rather than as a unique correct human reasoning chain. Based on this representation, we evaluate the functional quality of each stage and compare the decision value of the two types of process variables. We also examine whether the model-generated stages form a mutually supportive review chain and how intermediate review evidence relates to the final decision.
\subsection{Process-Aligned Data Construction}
We construct process-aligned data from PeerRead,
NLPeer ARR-22, and OpenReview-ICLR. These datasets cover an early peer-review corpus, the ACL Rolling Review process, and public ICLR review records, respectively, and differ in field structure, text granularity, and label sources. We use field-aware alignment: \(z_s\) uses reviewer-written summaries when available and paper abstracts otherwise; \(z_c\) and \(z_r\) are constructed from weakness-related and suggestion-related fields, respectively, with rule-based extraction from review text when structured fields are missing. The aligned \((z_s,z_c,z_r)\) form the Gold-process variables \(Z_G\). This conversion is more than text cleaning: it creates a shared process-aligned representation in which process variables can be independently evaluated, compared, and intervened on under a common protocol.

To enable cross-dataset comparison, we normalize decisions to
accept/reject. PeerRead uses original labels and OpenReview-ICLR uses
official decision notes. NLPeer ARR-22 lacks stable official binary
acceptance decisions, so we construct a score-based proxy decision using
an average review-score threshold of 3.5, which lies near the midpoint of
the rating scale. We filter samples without valid paper text or decision
labels; when a particular process variable is unavailable, the sample is
excluded only from evaluations that require that variable. The resulting
datasets contain 562, 364, and 12,002 samples, respectively. We use the
full test splits of PeerRead and NLPeer. For OpenReview-ICLR, we draw 500
label-stratified test samples per seed to control multi-stage inference
cost while approximately preserving the original label distribution.
Complete field mappings, filtering rules, and sensitivity results are
provided in Appendix A. 
\subsection{Evaluation Tasks and Stages}
As shown in Figure 1, we represent peer review as four stages: summary, critique, suggestion, and decision. Our six-stage protocol evaluates them using the Gold-process variables \(Z_G\) and Predicted- process variables \(\hat Z_P\) from Section 3.1.

\textbf{Stage 1: Direct Tasks.} This stage evaluates the model’s ability to perform basic review tasks and establishes a Direct baseline without intermediate process variables. The model performs two independent tasks: generating a summary \(\hat{z}_s\) from the paper content \(x\), and directly predicting a decision \(\hat{y}^{\mathrm{direct}}\) from \(x\) alone.

\textbf{Stage 2: Gold-process Decision Ablation.} This stage compares the relative decision value of different Gold-process variables and their combinations. The model predicts the decision using the seven non-empty subsets of \(Z_G\). We refer to the setting that uses the complete \(Z_G\) alone as Gold-only. We also include Paper + Gold, which jointly inputs the paper \(x\) and the complete \(Z_G\). Comparing it with Gold-only measures the effect of adding paper content.

\textbf{Stage 3: Predicted-process Generation and Decision.} This stage evaluates whether model-generated process variables can support the final decision. The model independently generates \(\hat{z}_s\) and \(\hat{z}_c\) from the paper \(x\), and then generates \(\hat{z}_r\) from \((x,\hat{z}_c)\). It predicts the decision using the seven non-empty subsets of \(\hat{Z}_P\). We refer to the setting that uses the complete \(\hat{Z}_P\) alone as Pred-only. Paper + Pred jointly inputs \((x,\hat{Z}_P)\) to test whether Predicted-process variables provide additional information when the paper content is available.

Stages 2 and 3 use the same seven process-variable subsets. We compare the decision performance of each corresponding pair to measure the decision-value gap between Gold-process variables and Predicted-process variables.

Stages 4–6 focus on Predicted-process variables. Their goal is to diagnose how the model generates and uses its own review evidence. Gold-process variables mainly serve as reference representations for comparing decision value, so they are not included in the subsequent process diagnosis.

\textbf{Stage 4: Chain-consistency Evaluation.} This stage evaluates whether the model-generated stage outputs form a mutually supportive review chain. Summary–critique consistency (S–C) examines whether the predicted critique logically follows from, or is at least compatible with, the predicted summary. Critique--suggestion consistency (C--S) checks whether the predicted suggestion addresses the critique; process--decision consistency (P--D) checks whether the critique and suggestion support the final decision.

\textbf{Stage 5: Interventional Sensitivity Analysis.} This stage tests whether the final decision responds to controlled changes in model-generated process variables. We intervene on only one variable, \(\hat{z}_s\), \(\hat{z}_c\), or \(\hat{z}_r\), at a time while keeping the other process variables and the decision model unchanged. We use four intervention types: positive, negative, neutral, and remove. The first three preserve the original process text and add a statement that supports acceptance, supports rejection, or has no clear decision preference. The remove intervention replaces the target variable with an explicit missing-value placeholder. We then repredict decisions under Pred-only (process-only) and
Paper + Pred (paper-conditioned) and compare them with their baselines.

\textbf{Stage 6: Conditional Error Analysis.} This stage examines whether final decision performance decreases when intermediate review stages fail to meet predefined quality conditions. For summary, critique, and suggestion, we divide samples into two groups according to the stage-level metrics: those that meet the quality condition and those that do not. We compare decision accuracy between the two groups under both process-only and paper-conditioned settings. The accuracy gap measures a potential error-propagation relationship, but it represents only a statistical association and does not provide causal evidence. Appendices B and C detail all process-variable combinations, inputs, outputs, prompts, intervention templates, and quality conditions.
\subsection{Evaluation Metrics}
We use task-specific metrics to evaluate stage-level quality, process variable decision value, chain consistency, interventional sensitivity, and conditional error analysis. Because these metrics measure different functions on different scales, we report them separately.

\textbf{Stage-level Metrics.} We use metrics that match the function of each review stage. For summary, ROUGE-1/2/L~\cite{lin2004rouge} and BERTScore-F1~\cite{zhang2020bertscore} measure lexical overlap and semantic similarity with the reference summary.

A critique may contain several distinct weaknesses. We therefore split generated and reference critiques into weakness units, then greedily match them one-to-one using unigram/bigram TF--IDF cosine similarity~\cite{salton1988term}. Matches require similarity of at least \(\tau_c=0.25\). Weakness Precision, Recall, and F1 measure matching from the predicted and reference sides; Coverage is the fraction of reference weakness units that are matched. This unit-level protocol identifies whether the model covers the specific issues raised in reference reviews, rather than relying on similarity between entire critique texts.

For suggestion, direct matching with a reference suggestion is not required because one weakness can admit multiple valid suggestions. Critique--Suggestion Alignment is the fraction of predicted critique units matched by at least one generated suggestion, using \(\tau_r=0.20\). A fixed LLM Judge~\cite{liu2023geval,zheng2023judging} additionally evaluates the paper, predicted critique, and generated suggestion for relevance, specificity, actionability, and overall quality on a 1--5 scale, where higher scores are better. For decision, we use Macro-F1 as the primary metric and also report Accuracy and ROC-AUC \cite{sokolova2009systematic,fawcett2006roc}. The matching thresholds define operational matching rules rather than universal quality cutoffs; implementation details and sensitivity analyses are provided in Appendix C.

\textbf{Process-variable Decision Value.} To compare the reference and model-generated representations of the same process-variable subset, let  $A \subseteq \{s,c,r\}$ denote any non-empty subset. We define $\Delta_A = M(\hat{y}_A^{\mathrm{gold}},y) - M(\hat{y}_A^{\mathrm{pred}},y),$ where \(M\) denotes Macro-F1. If \(\Delta_A>0\), the corresponding Gold-process subset has higher decision value. For the complete process, we denote the gap as $\Delta_{\mathrm{process}}=\Delta_{\{s,c,r\}}.$ 

\textbf{Chain Consistency.} A fixed LLM Judge evaluates three relations: summary–critique (S–C), critique–suggestion (C–S), and process–decision (P–D). Each relation receives a score from 1 to 5. A higher score indicates stronger support across stages. Overall CC is the arithmetic mean of the three scores. These metrics measure support between stages rather than the independent quality of individual outputs. Appendix F provides the complete scoring rubric.

\textbf{Interventional Sensitivity.} Let \(p_i\) and \(\hat{y}_i\) denote the acceptance probability and decision for sample \(i\) before intervention. Let \(p_i^{(k,t)}\) and \(\hat{y}_i^{(k,t)}\) denote the corresponding results after applying intervention type \(t\) to process variable \(k\). We compute $\Delta P_{k,t} = \frac{1}{N} \sum_{i=1}^{N} \left( p_i^{(k,t)}-p_i \right) $ and $\mathrm{DCR}_{k,t} = \frac{1}{N} \sum_{i=1}^{N} \mathbb{I} \left[ \hat{y}_i^{(k,t)}\neq\hat{y}_i \right].$
Here, $\Delta P$ measures the mean acceptance-probability shift, and DCR
the fraction of labels changed by intervention. We compute both under
process-only and paper-conditioned settings.

\textbf{Conditional Error Analysis.} We define the error-propagation metric as $\mathrm{EP}_{k\rightarrow y} = \operatorname{Acc}(\hat{y},y\mid e_k=0) - \operatorname{Acc}(\hat{y},y\mid e_k=1),$ where \(e_k=0\) and \(e_k=1\) indicate that stage \(k\) meets or does not meet the predefined quality condition, respectively. A positive value indicates that lower stage quality is associated with lower final decision accuracy. This metric measures only a statistical association and does not provide causal evidence. Appendices C, F, G, and H provide the complete threshold sensitivity analyses, Judge rubrics, intervention templates, and conditional-group statistics.
\section{Experiment}
This section reports results for the experimental stages defined in Section 3.3 and further examines whether the main findings generalize across models and datasets.
\subsection{Experimental setup}
We evaluate six models: Qwen2.5-7B-Instruct, Qwen2.5-14B-Instruct,
Llama-3.1-8B-Instruct, Llama-3.3-70B, GPT-4.1, and DeepReviewer-7B.
They cover general-purpose open models of different scales, a proprietary
API model, and a peer-review-oriented model. This selection tests whether
the main findings generalize across model scale, access mode, and domain
adaptation. Qwen2.5-14B-Instruct is the main analysis model because it can reproducibly run the complete six-stage protocol under the shared 32K
input budget. We run it on all datasets with seeds 42, 123, and 2024. The remaining five models run Stages 1--3 on the seed-42 splits of
PeerRead and NLPeer ARR-22 to compare process-variable decision value
across models. All models use shared prompts, a 32K full-if-fit plus section-aware fallback input policy, and deterministic decoding (temperature \(=0\)). Main-model results are reported as mean \(\pm\) sample SD over three seeds. Model access, full run coverage, generation lengths, and output protocols are provided in Appendix B.
\subsection{Process-variable Decision Value: Gold versus Predicted Process}
This section jointly analyzes Stages 2 and 3. We examine whether Gold-process variables can support the final decision, whether the model can generate Predicted-process variables with decision value, and how large the gap between them remains. Table 1 compares Direct, the seven corresponding Gold-process and Predicted-process subsets, and the paper-conditioned settings for the complete process. It reports decision Macro-F1 as Mean \(\pm\) Std over three seeds, together with \(\Delta_A\). Appendix D reports Accuracy, class-wise F1, AUC, per-seed results, and prediction distributions.
\begin{table*}[t]
\centering
\scriptsize
\setlength{\tabcolsep}{2.8pt}
\label{tab:process_decision_value}
\begin{tabular}{@{}lccccccccc@{}}
\toprule
& \multicolumn{3}{c}{PeerRead}
& \multicolumn{3}{c}{NLPeer ARR-22}
& \multicolumn{3}{c}{OpenReview-ICLR} \\
\cmidrule(lr){2-4}
\cmidrule(lr){5-7}
\cmidrule(lr){8-10}
Input
& G & P & $\Delta_A$
& G & P & $\Delta_A$
& G & P & $\Delta_A$ \\
\midrule

$x$
& \multicolumn{3}{c}{0.235$\pm$0.000}
& \multicolumn{3}{c}{0.347$\pm$0.001}
& \multicolumn{3}{c}{0.281$\pm$0.000} \\

$z_s$
& 0.282$\pm$0.010 & 0.238$\pm$0.000 & 0.044$\pm$0.010
& 0.461$\pm$0.007 & 0.347$\pm$0.000 & 0.114$\pm$0.007
& 0.338$\pm$0.012 & 0.284$\pm$0.004 & 0.054$\pm$0.009 \\

$z_c$
& 0.637$\pm$0.008 & 0.409$\pm$0.000 & 0.227$\pm$0.008
& 0.472$\pm$0.012 & 0.319$\pm$0.000 & 0.153$\pm$0.012
& 0.551$\pm$0.029 & 0.379$\pm$0.000 & 0.172$\pm$0.029 \\

$z_r$
& 0.639$\pm$0.011 & 0.455$\pm$0.008 & 0.184$\pm$0.004
& 0.545$\pm$0.014 & 0.511$\pm$0.014 & 0.034$\pm$0.028
& 0.661$\pm$0.008 & 0.481$\pm$0.006 & 0.181$\pm$0.013 \\

$(z_s,z_c)$
& 0.663$\pm$0.001 & 0.466$\pm$0.016 & 0.197$\pm$0.016
& 0.677$\pm$0.007 & 0.444$\pm$0.022 & 0.233$\pm$0.019
& 0.711$\pm$0.019 & 0.478$\pm$0.035 & 0.233$\pm$0.028 \\

$(z_s,z_r)$
& 0.514$\pm$0.007 & 0.240$\pm$0.002 & 0.274$\pm$0.009
& 0.527$\pm$0.028 & 0.347$\pm$0.000 & 0.180$\pm$0.028
& 0.594$\pm$0.014 & 0.281$\pm$0.000 & 0.313$\pm$0.014 \\

$(z_c,z_r)$
& \textbf{0.728$\pm$0.014} & 0.409$\pm$0.000 & 0.319$\pm$0.014
& 0.505$\pm$0.007 & 0.319$\pm$0.000 & 0.187$\pm$0.007
& 0.589$\pm$0.023 & 0.379$\pm$0.000 & 0.210$\pm$0.023 \\

$Z_G/\hat{Z}_P$
& 0.649$\pm$0.005 & \textbf{0.504$\pm$0.013} & 0.145$\pm$0.017
& \textbf{0.683$\pm$0.005} & \textbf{0.541$\pm$0.012} & 0.142$\pm$0.014
& \textbf{0.715$\pm$0.015} & \textbf{0.556$\pm$0.021} & 0.159$\pm$0.016 \\

$(x,Z_G)/(x,\hat{Z}_P)$
& 0.491$\pm$0.001 & 0.238$\pm$0.000 & 0.253$\pm$0.001
& 0.513$\pm$0.011 & 0.347$\pm$0.003 & 0.166$\pm$0.014
& 0.627$\pm$0.024 & 0.291$\pm$0.005 & 0.336$\pm$0.027 \\

\bottomrule
\end{tabular}
\caption{Decision Macro-F1 across input settings (mean $\pm$ sample SD over three seeds). G and P denote Gold-process and Predicted-process results; $\Delta_A$ is the paired per-seed difference $\mathrm{G}-\mathrm{P}$. Bold marks the best G and P results per dataset.}
\end{table*}

\textbf{Gold-process variables have higher decision value, but a gap remains between Gold-process and Predicted-process variables.} Gold-process outperforms the corresponding Predicted-process in all 21 subset–dataset comparisons. Complete Gold-only improves over Direct by 0.336–0.434. Pred-only also improves over Direct by 0.194–0.275, but remains 0.142–0.159 below Gold-only. The best variable combination differs across datasets.Critique + Suggestion leads on PeerRead, while the complete process
leads on NLPeer ARR-22 and OpenReview-ICLR.

\textbf{Joint input with the paper does not ensure effective use of process variables.} Paper + Gold outperforms Direct on all three datasets. In contrast, Paper + Pred improves over Direct by no more than 0.010 and performs substantially worse than Pred-only. These results show that Predicted-process variables have some decision value but are not consistently used in the current joint-input setting.
\subsection{Stage-level Evaluation}
This section evaluates the local performance of the model at four review stages: summary, critique, suggestion, and decision. It also identifies possible sources of the Gold–Predicted gap observed in Section 4.2. Table 2 reports Mean \(\pm\) Std results for Qwen2.5-14B-Instruct over three seeds. Appendix E provides the complete per-seed results.
\begin{table*}[t]
\centering
\footnotesize
\setlength{\tabcolsep}{2.0pt}
\renewcommand{\arraystretch}{1.08}

\newcommand{\mstd}[2]{%
  \shortstack{$#1$\\[-1pt]{\scriptsize$\pm #2$}}%
}
\label{tab:stage_level_results}

\begin{tabular}{@{}lccccccccc@{}}
\toprule
& \multicolumn{2}{c}{Summary}
& \multicolumn{2}{c}{Critique}
& \multicolumn{2}{c}{Suggestion}
& \multicolumn{3}{c}{Decision} \\
\cmidrule(lr){2-3}
\cmidrule(lr){4-5}
\cmidrule(lr){6-7}
\cmidrule(lr){8-10}

Dataset
& \shortstack{ROUGE-L}
& \shortstack{BERT-F1}
& \shortstack{Weak.-F1}
& \shortstack{Coverage}
& \shortstack{C--S Align.}
& \shortstack{Judge Overall}
& \shortstack{Accuracy}
& \shortstack{Macro-F1}
& \shortstack{ROC-AUC} \\
\midrule

PeerRead
& \mstd{0.322}{0.002}
& \mstd{0.890}{0.001}
& \mstd{0.0011}{0.0003}
& \mstd{0.0023}{0.0016}
& \mstd{0.484}{0.006}
& \mstd{4.374}{0.014}
& \mstd{0.307}{0.000}
& \mstd{0.235}{0.000}
& \mstd{0.580}{0.003} \\

NLPeer ARR-22
& \mstd{0.214}{0.002}
& \mstd{0.866}{0.001}
& \mstd{0.0033}{0.0009}
& \mstd{0.0036}{0.0009}
& \mstd{0.495}{0.006}
& \mstd{4.336}{0.017}
& \mstd{0.530}{0.002}
& \mstd{0.347}{0.001}
& \mstd{0.504}{0.011} \\

OpenReview-ICLR
& \mstd{0.234}{0.005}
& \mstd{0.870}{0.000}
& \mstd{0.0040}{0.0002}
& \mstd{0.0063}{0.0033}
& \mstd{0.484}{0.006}
& \mstd{4.493}{0.013}
& \mstd{0.390}{0.000}
& \mstd{0.281}{0.000}
& \mstd{0.465}{0.002} \\

\bottomrule
\end{tabular}
\caption{Stage-level results for Qwen2.5-14B-Instruct (mean $\pm$ sample SD over three seeds). BERT-F1, Weak.-F1, and C--S Align. denote BERTScore-F1, Weakness F1, and Critique--Suggestion Alignment. Judge Overall uses a 1--5 scale; all other metrics use a 0--1 scale.}
\end{table*}

\textbf{Performance varies substantially across review stages.} Summary BERTScore-F1 is similar across the three datasets (0.866–0.890). In contrast, predicted critiques show very low strict alignment with reference weaknesses, with both Weakness F1 and Coverage close to zero. This does not imply that the generated critiques are entirely invalid. Rather, identified issues rarely match specific reference weaknesses under this protocol, which may contribute to the Gold--Predicted gap.

\textbf{High local suggestion quality does not translate into reliable decisions.} The LLM Judge assigns suggestion Overall scores of 4.336–4.493, whereas Critique–Suggestion Alignment remains only 0.484–0.495. Direct Macro-F1 is also low (0.235–0.347) and shows a clear accept bias. Overall, the model can generate locally plausible review text, but still struggles to align critiques with reference weaknesses, respond comprehensively to preceding critiques, and distinguish between accept and reject decisions. Next, we test whether they form a mutually supportive review chain.
\subsection{Chain-consistency Evaluation}
This section examines whether the stage outputs in the Predicted-process form a coherent and mutually supportive review process. Table 3 reports the Judge's raw scores on a 1--5 scale as mean $\pm$ sample standard deviation over three seeds. The complete scoring rubric, numbers of valid samples, and per-seed results are provided in Appendix F.
\begin{table}[t]
\centering
\label{tab:chain_consistency}
\footnotesize
\setlength{\tabcolsep}{2.6pt}
\begin{tabular}{@{}lcccc@{}}
\toprule
Dataset
& S--C
& C--S
& P--D
& \shortstack{Overall\\CC} \\
\midrule
PeerRead
& 3.986$\pm$0.003
& 4.999$\pm$0.001
& 1.312$\pm$0.022
& 3.432$\pm$0.008 \\

NLPeer
& 4.003$\pm$0.010
& 4.999$\pm$0.002
& 1.325$\pm$0.008
& 3.442$\pm$0.003 \\

OpenReview
& 3.997$\pm$0.012
& 4.989$\pm$0.008
& 1.409$\pm$0.011
& 3.465$\pm$0.004 \\
\bottomrule
\end{tabular}
\caption{Predicted-process chain consistency (mean $\pm$ sample SD over three seeds). Fixed-judge scores range from 1 to 5; Overall CC averages S--C, C--S, and P--D.}
\end{table}
\textbf{The final decision is the main break in the review chain.} S--C scores are close to 4.0 and C--S scores are close to 5.0 across all three datasets. In contrast, P--D scores are only 1.31--1.41, with 97.0\%--97.9\% of samples receiving scores of 1 or 2. These results show high local consistency in generated intermediate texts, but final decisions often lack support from preceding critiques and suggestions.

\textbf{Conditional error analysis does not reveal a consistent error-propagation pattern.} The conditional accuracy gaps for summary are small and vary in direction, while the estimates for critique and suggestion are affected by severe imbalance between the condition groups. These results mainly reveal the limitations of converting stage quality into binary error labels and should not be interpreted as evidence of causal error propagation. The next section uses controlled interventions to examine more directly whether the final decision responds to changes in intermediate process variables.
\subsection{Interventional Sensitivity Analysis}
This section examines whether the final decision changes in response to controlled modifications of intermediate process variables. The main analysis uses the process-only setting. We repeat the same interventions under the paper-conditioned setting to examine whether the model still uses process evidence when the paper content is included. Both settings share templates, model, and metrics.
\begin{figure*}[!t]
\centering
\includegraphics[width=0.65\textwidth]{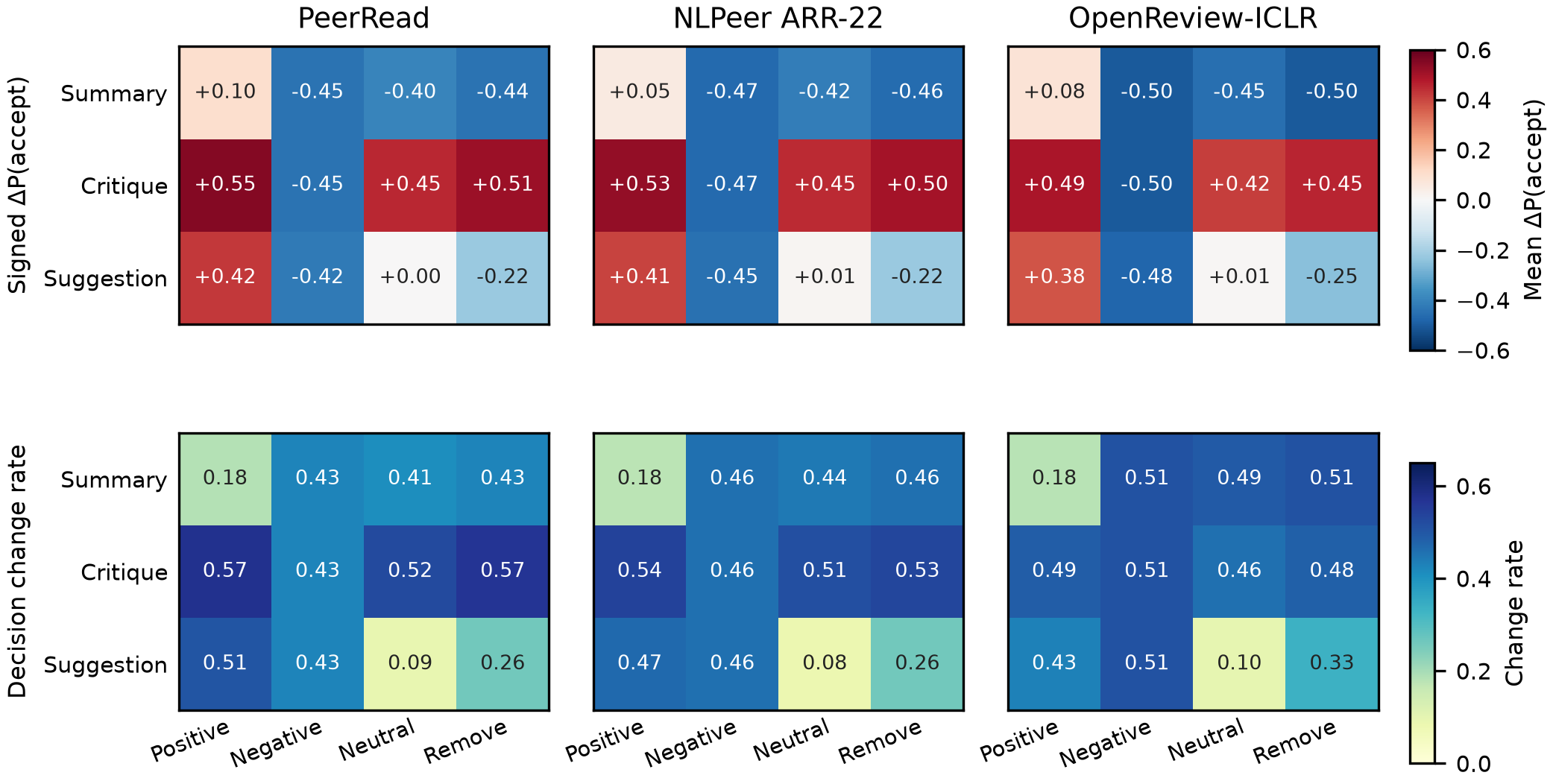} 
\caption{Process-only intervention effects on Predicted-process variables (three-seed means). Top: acceptance-probability change $\Delta P$; bottom: Decision Change Rate (DCR). Positive and negative $\Delta P$ favor accept and reject, respectively.}
\label{fig:results}
\end{figure*} 
\begin{table*}[!t]
\centering
\small
\setlength{\tabcolsep}{4.2pt}
\begin{tabular*}{\textwidth}{
@{\extracolsep{\fill}}
lcccccccc
@{}
}
\toprule
& \multicolumn{4}{c}{PeerRead}
& \multicolumn{4}{c}{NLPeer ARR-22} \\
\cmidrule(lr){2-5}
\cmidrule(lr){6-9}
Model & D & G & P & Gap & D & G & P & Gap \\
\midrule
Qwen2.5-7B
& 0.233 & \textbf{0.633} & 0.419 & 0.214
& 0.373 & \textbf{0.578} & 0.338 & 0.240 \\

Qwen2.5-14B
& 0.235 & \textbf{0.644} & 0.515 & 0.129
& 0.346 & \textbf{0.677} & 0.539 & 0.138 \\

Llama-3.1-8B
& 0.237 & \textbf{0.570} & 0.499 & 0.071
& 0.354 & \textbf{0.704} & 0.459 & 0.245 \\

Llama-3.3-70B
& 0.235 & \textbf{0.641} & 0.526 & 0.115
& 0.348 & \textbf{0.708} & 0.483 & 0.225 \\

GPT-4.1
& 0.245 & \textbf{0.674} & 0.409 & 0.265
& 0.346 & \textbf{0.679} & 0.319 & 0.360 \\

DeepReviewer-7B
& 0.281 & 0.350 & \textbf{0.356} & -0.006
& 0.509 & \textbf{0.510} & 0.427 & 0.083 \\
\bottomrule
\end{tabular*}
\caption{Cross-model decision Macro-F1 on the seed-42 test splits. D, G, and P denote Direct, Gold-process, and Predicted-process; Gap is $\mathrm{G}-\mathrm{P}$. Bold marks the best setting per model--dataset pair.}
\label{tab:cross_model}
\end{table*}

\textbf{The final decision responds to controlled changes in process variables, with critique having the largest effect.} As shown in Figure 3, positive and negative interventions produce the expected directions of \(\Delta P\) across all three datasets. Critique has an average \(|\Delta P|\) of 0.468--0.491 and a DCR of 0.485--0.523, both higher than those of summary and suggestion. This shows that the final decision is most sensitive to explicit evaluative evidence in the critique.

\textbf{Including the paper weakens the effects of most process interventions.} Under the paper-conditioned setting, most interventions have substantially smaller effects. However, negative critique can still overturn the original accept bias, with a DCR of 0.987--0.997 and a decrease of 0.927--0.963 in acceptance probability. This asymmetric response shows that the model reacts to changes in process evidence but does not yet use such evidence in a balanced and stable manner. Full process-only and paper-conditioned results are in Appendix G. These results indicate sensitivity, not causality.

\subsection{Model and Dataset Generalization}
This section examines whether the preceding findings generalize across models and datasets. We compare Qwen2.5-7B, Qwen2.5-14B, Llama-3.1-8B, Llama-3.3-70B, GPT-4.1, and DeepReviewer-7B on the same seed-42 test splits of PeerRead and NLPeer ARR-22. All models use the same process-variable definitions, input structure, and decision-evaluation protocol. Table 4 reports Macro-F1 for the Direct, Gold-process, and Predicted-process settings. Appendix I reports additional metrics and prediction distributions; Appendix D reports all main-model results across datasets and seeds.

\textbf{The decision advantage of Gold-process variables generalizes across general-purpose models.} As shown in Table 4, Gold-process outperforms Predicted-process in all ten model--dataset combinations formed by five general-purpose models and two datasets. The gaps range from 0.071 to 0.265 on PeerRead and from 0.137 to 0.360 on NLPeer ARR-22. Clear gaps remain for Llama-3.3-70B and GPT-4.1, showing that larger models do not consistently close the gap.

\textbf{A smaller gap does not necessarily indicate stronger process generation.} On PeerRead, DeepReviewer-7B obtains similar Gold-process and Predicted-process results, but both Macro-F1 scores are approximately 0.35. Its small gap therefore results mainly from weak Gold-process performance. The improvement of Predicted-process over Direct also varies across models and datasets. Overall, the higher decision value of Gold-process variables generalizes across models, whereas the ability to generate and use effective process variables remains inconsistent.

\section{Conclusion}
We introduce a process-centric diagnostic benchmark for AI-assisted peer review. It aligns paper content, summary, critique, suggestion, and decision into a unified process-aligned representation and evaluates stage-level quality, process-variable decision value, chain consistency, and interventional sensitivity. Experiments across datasets and models show that Gold-process variables have higher decision value than Predicted-process variables. Although model-generated review texts show high local consistency across adjacent stages, final decisions are not consistently supported by preceding review evidence. 

Our benchmark is intended to support, rather than replace, human reviewers. It provides a transparent and auditable diagnostic tool for identifying deficiencies in review evidence generation and use. Future work will incorporate expert validation and further examine bias and uncertainty to support the safe and reliable use of AI in peer review under human oversight.

\bigskip

\bibliography{aaai2027}

@inproceedings{kang2018dataset,
  title={A dataset of peer reviews (PeerRead): Collection, insights and NLP applications},
  author={Kang, Dongyeop and Ammar, Waleed and Dalvi, Bhavana and Van Zuylen, Madeleine and Kohlmeier, Sebastian and Hovy, Eduard and Schwartz, Roy},
  booktitle={Proceedings of the 2018 conference of the North American chapter of the Association for Computational Linguistics: Human language technologies, volume 1 (long papers)},
  pages={1647--1661},
  year={2018}
}

@article{liang2023can,
  title={Can large language models provide useful feedback on research papers? A large-scale empirical analysis. arXiv},
  author={Liang, W and Zhang, Y and Cao, H and Wang, B and Ding, D and Yang, X and Vodrahalli, K and He, S and Smith, D and Yin, Y and others},
  journal={arXiv preprint arXiv:2310.01783},
  year={2023}
}

@inproceedings{idahl2025openreviewer,
  title={Openreviewer: A specialized large language model for generating critical scientific paper reviews},
  author={Idahl, Maximilian and Ahmadi, Zahra},
  booktitle={Proceedings of the 2025 Conference of the Nations of the Americas Chapter of the Association for Computational Linguistics: Human Language Technologies (System Demonstrations)},
  pages={550--562},
  year={2025}
}

@article{biswas2026ai,
  title={AI-assisted peer review at scale: The AAAI-26 AI review pilot},
  author={Biswas, Joydeep and Schoepp, Sheila and Vasan, Gautham and Opipari, Anthony and Zhang, Arthur and Hu, Zichao and Joseph, Sebastian and Lease, Matthew and Li, Junyi Jessy and Stone, Peter and others},
  journal={arXiv preprint arXiv:2604.13940},
  year={2026}
}

@inproceedings{zhou2024llm,
  title={Is LLM a reliable reviewer? A comprehensive evaluation of LLM on automatic paper reviewing tasks},
  author={Zhou, Ruiyang and Chen, Lu and Yu, Kai},
  booktitle={Proceedings of the 2024 joint international conference on computational linguistics, language resources and evaluation (LREC-COLING 2024)},
  pages={9340--9351},
  year={2024}
}

@article{d2024marg,
  title={Marg: Multi-agent review generation for scientific papers},
  author={D'Arcy, Mike and Hope, Tom and Birnbaum, Larry and Downey, Doug},
  journal={arXiv preprint arXiv:2401.04259},
  year={2024}
}

@inproceedings{weng2025cycleresearcher,
  title={Cycleresearcher: Improving automated research via automated review},
  author={Weng, Yixuan and Zhu, Minjun and Bao, Guangsheng and Zhang, Hongbo and Wang, Jindong and Zhang, Yue and Yang, Linyi},
  booktitle={International Conference on Learning Representations},
  volume={2025},
  pages={3669--3709},
  year={2025}
}

@inproceedings{zhu2025deepreview,
  title={Deepreview: Improving llm-based paper review with human-like deep thinking process},
  author={Zhu, Minjun and Weng, Yixuan and Yang, Linyi and Zhang, Yue},
  booktitle={Proceedings of the 63rd Annual Meeting of the Association for Computational Linguistics (Volume 1: Long Papers)},
  pages={29330--29355},
  year={2025}
}

@article{garg2025revieweval,
  title={Revieweval: An evaluation framework for ai-generated reviews},
  author={Garg, Madhav Krishan and Prasad, Tejash and Singhal, Tanmay and Kirtani, Chhavi and Mandal, Murari and Kumar, Dhruv},
  journal={arXiv preprint arXiv:2502.11736},
  year={2025}
}

@inproceedings{dycke2023nlpeer,
  title={NLPeer: A unified resource for the computational study of peer review},
  author={Dycke, Nils and Kuznetsov, Ilia and Gurevych, Iryna},
  booktitle={Proceedings of the 61st annual meeting of the Association for Computational Linguistics (volume 1: Long papers)},
  pages={5049--5073},
  year={2023}
}

@article{wang2023have,
  title={What have we learned from OpenReview?},
  author={Wang, Gang and Peng, Qi and Zhang, Yanfeng and Zhang, Mingyang},
  journal={World Wide Web},
  volume={26},
  number={2},
  pages={683--708},
  year={2023},
  publisher={Springer}
}

@article{yuan2022can,
  title={Can we automate scientific reviewing?},
  author={Yuan, Weizhe and Liu, Pengfei and Neubig, Graham},
  journal={Journal of Artificial Intelligence Research},
  volume={75},
  pages={171--212},
  year={2022}
}

@inproceedings{chang2025treereview,
  title={TreeReview: A dynamic tree of questions framework for deep and efficient LLM-based scientific peer review},
  author={Chang, Yuan and Li, Ziyue and Zhang, Hengyuan and Kong, Yuanbo and Wu, Yanru and So, Hayden Kwok-Hay and Guo, Zhijiang and Zhu, Liya and Wong, Ngai},
  booktitle={Proceedings of the 2025 Conference on Empirical Methods in Natural Language Processing},
  pages={15662--15693},
  year={2025}
}

@article{goyal2026scholarpeer,
  title={ScholarPeer: A Context-Aware Multi-Agent Framework for Automated Peer Review},
  author={Goyal, Palash and Parmar, Mihir and Song, Yiwen and Palangi, Hamid and Pfister, Tomas and Yoon, Jinsung},
  journal={arXiv preprint arXiv:2601.22638},
  year={2026}
}

@misc{sahu2510reviewertoo,
  title         = {ReviewerToo: Should {AI} Join the Program Committee?
                   A Look at the Future of Peer Review},
  author        = {Sahu, Gaurav and Larochelle, Hugo and Charlin, Laurent
                   and Pal, Christopher},
  year          = {2025},
  eprint        = {2510.08867},
  archivePrefix = {arXiv},
  url           = {https://arxiv.org/abs/2510.08867}
}

@article{li2026beyond,
  title={Beyond Rating: A Comprehensive Evaluation and Benchmark for AI Reviews},
  author={Li, Bowen and Ma, Haochen and Wang, Yuxin and Yang, Jie and Zheng, Yining and Chen, Xinchi and Huang, Xuanjing and Qiu, Xipeng},
  journal={arXiv preprint arXiv:2604.19502},
  year={2026}
}

@article{gao2025mmreview,
  title={MMReview: A Multidisciplinary and Multimodal Benchmark for LLM-Based Peer Review Automation},
  author={Gao, Xian and Ruan, Jiacheng and Zhang, Zongyun and Gao, Jingsheng and Liu, Ting and Fu, Yuzhuo},
  journal={arXiv preprint arXiv:2508.14146},
  year={2025}
}

@article{loc2026prism,
  title={PRISM: A Multi-Dimensional Benchmark for Evaluating LLM Peer Reviewers},
  author={Loc, Ngoc Phan Phuoc and Viet, La and Huynh, Toan and Khanh, Thanh Tran and Nguyen, Duy A and Pham, Tuan Anh Nguyen and Nguyen, Thanh and Chawla, Nitesh V and Buntine, Wray and Wong, Kok-Seng and others},
  journal={arXiv preprint arXiv:2605.26730},
  year={2026}
}

@article{deng2026cocoreviewbench,
  title={CoCoReviewBench: A Completeness-and Correctness-Oriented Benchmark for AI Reviewers},
  author={Deng, Hexuan and Ke, Xiaopeng and Li, Yichen and Hu, Ruina and Huang, Dehao and Wong, Derek F and Wang, Yue and Liu, Xuebo and Zhang, Min},
  journal={arXiv preprint arXiv:2605.07905},
  year={2026}
}

@article{jin2026makes,
  title={What Makes a Good AI Review? Concern-Level Diagnostics for AI Peer Review},
  author={Jin, Ming},
  journal={arXiv preprint arXiv:2604.19998},
  year={2026}
}

@article{li2022peersum,
  title={Peersum: a peer review dataset for abstractive multi-document summarization},
  author={Li, Miao and Qi, Jianzhong and Lau, Jey Han},
  journal={arXiv preprint arXiv:2203.01769},
  year={2022}
}

@article{lin2023moprd,
  title={Moprd: A multidisciplinary open peer review dataset},
  author={Lin, Jialiang and Song, Jiaxin and Zhou, Zhangping and Chen, Yidong and Shi, Xiaodong},
  journal={Neural Computing and Applications},
  volume={35},
  number={34},
  pages={24191--24206},
  year={2023},
  publisher={Springer}
}

@inproceedings{fromm2021argument,
  title={Argument mining driven analysis of peer-reviews},
  author={Fromm, Michael and Faerman, Evgeniy and Berrendorf, Max and Bhargava, Siddharth and Qi, Ruoxia and Zhang, Yao and Dennert, Lukas and Selle, Sophia and Mao, Yang and Seidl, Thomas},
  booktitle={Proceedings of the AAAI conference on artificial intelligence},
  volume={35},
  pages={4758--4766},
  year={2021}
}

@inproceedings{lin2004rouge,
  title = {{ROUGE}: A Package for Automatic Evaluation of Summaries},
  author = {Lin, Chin-Yew},
  booktitle = {Text Summarization Branches Out},
  pages = {74--81},
  year = {2004},
  url = {https://aclanthology.org/W04-1013/}
}

@inproceedings{zhang2020bertscore,
  title = {{BERTScore}: Evaluating Text Generation with {BERT}},
  author = {Zhang, Tianyi and Kishore, Varsha and Wu, Felix and
            Weinberger, Kilian Q. and Artzi, Yoav},
  booktitle = {International Conference on Learning Representations},
  year = {2020},
  url = {https://openreview.net/forum?id=SkeHuCVFDr}
}

@article{salton1988term,
  title = {Term-Weighting Approaches in Automatic Text Retrieval},
  author = {Salton, Gerard and Buckley, Christopher},
  journal = {Information Processing \& Management},
  volume = {24},
  number = {5},
  pages = {513--523},
  year = {1988},
  doi = {10.1016/0306-4573(88)90021-0}
}

@inproceedings{liu2023geval,
  title = {{G}-{Eval}: {NLG} Evaluation using {GPT}-4 with Better Human Alignment},
  author = {Liu, Yang and Iter, Dan and Xu, Yichong and Wang, Shuohang and
            Xu, Ruochen and Zhu, Chenguang},
  booktitle = {Proceedings of the 2023 Conference on Empirical Methods in Natural Language Processing},
  pages = {2511--2522},
  year = {2023},
  url = {https://aclanthology.org/2023.emnlp-main.153/}
}

@inproceedings{zheng2023judging,
  title = {Judging {LLM}-as-a-Judge with {MT}-Bench and Chatbot Arena},
  author = {Zheng, Lianmin and Chiang, Wei-Lin and Sheng, Ying and
            Zhuang, Siyuan and Wu, Zhanghao and Zhuang, Yonghao and
            Lin, Zi and Li, Zhuohan and Li, Dacheng and Xing, Eric P. and
            Zhang, Hao and Gonzalez, Joseph E. and Stoica, Ion},
  booktitle = {Advances in Neural Information Processing Systems},
  volume = {36},
  year = {2023},
  url = {https://arxiv.org/abs/2306.05685}
}

@article{fawcett2006roc,
  title = {An Introduction to {ROC} Analysis},
  author = {Fawcett, Tom},
  journal = {Pattern Recognition Letters},
  volume = {27},
  number = {8},
  pages = {861--874},
  year = {2006},
  doi = {10.1016/j.patrec.2005.10.010}
}

@inproceedings{kim2025peerreviewcrisis,
  title     = {Position: The {AI} Conference Peer Review Crisis Demands
               Author Feedback and Reviewer Rewards},
  author    = {Kim, Jaeho and Lee, Yunseok and Lee, Seulki},
  booktitle = {Proceedings of the 42nd International Conference on
               Machine Learning},
  year      = {2025},
  url       = {https://openreview.net/forum?id=l8QemUZaIA}
}

@article{latona2024aireviewlottery,
  title   = {The {AI} Review Lottery: Widespread {AI}-Assisted Peer
             Reviews Boost Paper Scores and Acceptance Rates},
  author  = {Latona, Giuseppe Russo and Ribeiro, Manoel Horta and
             Davidson, Tim R. and Veselovsky, Veniamin and West, Robert},
  journal = {arXiv preprint arXiv:2405.02150},
  year    = {2024},
  doi     = {10.48550/arXiv.2405.02150},
  url     = {https://arxiv.org/abs/2405.02150}
}

@misc{aaai2025reviewpilot,
  title        = {{AI}-Assisted Peer-Review Process: Pilot Program},
  author       = {{Association for the Advancement of Artificial Intelligence}},
  year         = {2025},
  howpublished = {\url{https://aaai.org/conference/aaai/aaai-26/main-technical-track-call/}},
  note         = {AAAI-26 Main Technical Track; accessed July 25, 2026}
}

@misc{neurips2026reviewexperiment,
  title        = {{NeurIPS} 2026 {AI}-Assisted Reviewing Experiment},
  author       = {{NeurIPS}},
  year         = {2026},
  howpublished = {\url{https://dev.neurips.cc/Conferences/2026/ai-reviewing-experiment}},
  note         = {Accessed July 25, 2026}
}

@misc{neurips2025reviewerguidelines,
  title        = {{NeurIPS} 2025 Reviewer Guidelines},
  author       = {{Neural Information Processing Systems Foundation}},
  year         = {2025},
  howpublished = {\url{https://neurips.cc/Conferences/2025/ReviewerGuidelines}},
  note         = {Accessed July 25, 2026}
}

@article{sokolova2009systematic,
  title = {A Systematic Analysis of Performance Measures for Classification Tasks},
  author = {Sokolova, Marina and Lapalme, Guy},
  journal = {Information Processing \& Management},
  volume = {45},
  number = {4},
  pages = {427--437},
  year = {2009},
  publisher = {Elsevier},
  doi = {10.1016/j.ipm.2009.03.002},
  url = {https://doi.org/10.1016/j.ipm.2009.03.002}
}


\end{document}